\documentclass{article}
\usepackage[final]{corl_2020} 

\usepackage{graphicx} 
\usepackage{subfigure}
\usepackage{amssymb}
\usepackage{amsmath}
\usepackage{algorithm}
\usepackage{algpseudocode}
\usepackage{acronym}
\usepackage{capt-of}

\acrodef{ADR}{Active Domain Randomization}
\acrodef{RL}{Reinforcement Learning}
\acrodef{DR}{Domain Randomization}
\acrodef{UDR}{Uniform Domain Randomization}
\acrodef{DoF}{Degrees of Freedom}
\acrodef{MDP}{Markov Decision Process}
\usepackage{latexsym} 
\title{ Generating Automatic Curricula via Self-Supervised Active Domain Randomization}

\author{ 
{ Sharath Chandra Raparthy\thanks{~~Correspondence to \texttt{raparths@mila.quebec}}} \\
 Mila, Université de Montréal
\And
{ Bhairav Mehta} \\
Mila, Université de Montréal\\
\And
{ Florian Golemo}   \\
Mila, Université de Montréal \\
Element AI
\AND
{ Liam Paull}   \\
Université de Montréal, Mila \\
CIFAR AI Chair
}

\begin{document}
\maketitle

\begin{abstract}
Goal-directed \ac{RL} traditionally considers an agent interacting with an \textit{environment}, prescribing a real-valued reward to an agent proportional to the completion of some goal. Goal-directed \ac{RL} has seen large gains in sample efficiency, due to the ease of reusing or generating new experience by proposing goals. One approach, \textit{self-play}, allows an agent to ``play" against itself by alternatively setting and accomplishing goals, creating a learned curriculum through which an agent can learn to accomplish progressively more difficult goals. However, self-play has been limited to goal curriculum learning or learning progressively harder goals \textit{within} a single environment. Recent work on robotic agents has shown that varying the \textit{environment} during training, for example with domain randomization, leads to more robust transfer. As a result, we extend the self-play framework to jointly learn a goal and environment curriculum, leading to an approach that learns the most fruitful domain randomization strategy with self-play. Our method, \textit{Self-Supervised Active Domain Randomization} (SS-ADR), generates a coupled goal-task curriculum, where agents learn through progressively more difficult tasks and environment variations. By encouraging the agent to try tasks that are just outside of its current capabilities, SS-ADR builds a domain randomization curriculum that enables state-of-the-art results on various sim2real transfer tasks. Our results show that a curriculum of co-evolving the environment difficulty \textit{together} with the difficulty of goals set in each environment provides practical benefits in the goal-directed tasks tested.
\end{abstract}

\section{Introduction}

Reinforcement learning offers a way for autonomous agents to learn behaviors via interaction with an environment. A central, often practical, problem of reinforcement learning is \textit{reward engineering}: the process of creating reward functions that, when used during optimization, generate desirable behaviors. Reward design is an arduous, trial-and-error process, and provides experimenters little insight into how a particular reward generated certain behaviors. The reward design process stands in the way of reinforcement learning becoming a realistic contender for true artificial intelligence, with fundamental issues spawning entire new fields of study such as reward hacking \citep{Openaifaultyreward}, AI safety \citep{amodeiAiSafety}, and AI alignment \citep{gabriel2020artificial}. 

Reward design can also induce training difficulties, as certain reward functions may be too difficult for agents to optimize \citep{schaul2019ray}. Curriculum learning \citep{Bengio:2009:CL:1553374.1553380} seeks to ease the optimization process by allowing agents to solve progressively more difficult variants of a task. However, traditional curriculum learning often requires heuristics that control \textit{when} to advance agents between tasks, simply moving the reward design issue up a layer of abstraction.

An alternative to the standard reward design and curriculum learning paradigms is to have agents reward themselves. Numerous works explore the field of \textit{intrinsic motivation}, using metrics such as surprisal \citep{bartosuprisal, bellemare2016unifying}, Bayesian information gain \citep{shyam2018modelbased}, and curiosity \citep{burda2018largescale} to improve agent exploration. \textit{Automatic} curricula naturally arise in this scenario, as an agent is required to push its own frontiers in order to generate further intrinsic rewards. 

Self-play is a popular intrinsic motivation method that allows for training of agents without rewards, particularly in goal-directed reinforcement learning. Self-play pits two agents against each other in a standard \ac{MDP} framework. These two agents are often time-separated ``copies" of the same agent. Self-play rewards an agent for making progress against the other, generating a game-theoretic, automatic curriculum as each agent tries to get the upper hand.

While self-play has proved itself in non-equilibrium, two-player games \citep{Silver_2016, asymmetricselfplay}, one under explored application of self-play is in \textit{task selection} - or curriculum learning in the task, rather than goal, space. Many problems in reinforcement learning need not focus on such a problem, as task curriculum design is only relevant in transfer learning scenarios. 



One of the most exciting applications of transfer learning is the \textit{sim2real} problem in robotic learning. In robotic \ac{RL}, policies trained purely in a simulation have proved difficult to transfer to the real world, a problem known as \textit{``reality gap''} \citep{jakobi1995noise}. One leading approach for this sim2real transfer is \ac{DR} \citep{tobin2017domain}, where a simulator's parameters are perturbed, generating a space of similar yet distinct environments, all of which an agent tries to solve before transferring to a real robot. Similar to issues of \textit{which} goals to show an agent in goal curriculum learning, the issue once again becomes a question of \textit{which} environments to show the agent. Recently, \citet{pmlr-v100-mehta20a} empirically showed that not all generated environments are equally useful for learning, leading to \ac{ADR}. \ac{ADR} defines a curriculum learning problem in the environment randomized space, using \textit{learned rewards} to search for an optimal curriculum.

In this work, we propose the combination of \ac{ADR} and asymmetric self-play. We formulate our problem as a bilevel optimization where in the inner loop we maximize the expected return over the given environment-goal pairs, and in the outer loop, we maximize a self-play based metric to enforce a naturally growing environment-goal curriculum. We show that we can co-evolve curricula in \textit{both goal and environment spaces}, using only a single self-supervised reward. This bilevel formulation induces robustness in policy performance, further reducing variance compared to other curriculum learning methods. We show that this coupling generates strong robotic policies in all environments tested, even across multiple real world settings.

\begin{figure}[tb]
  \begin{minipage}[c]{0.5\textwidth}
    \includegraphics[width=\textwidth]{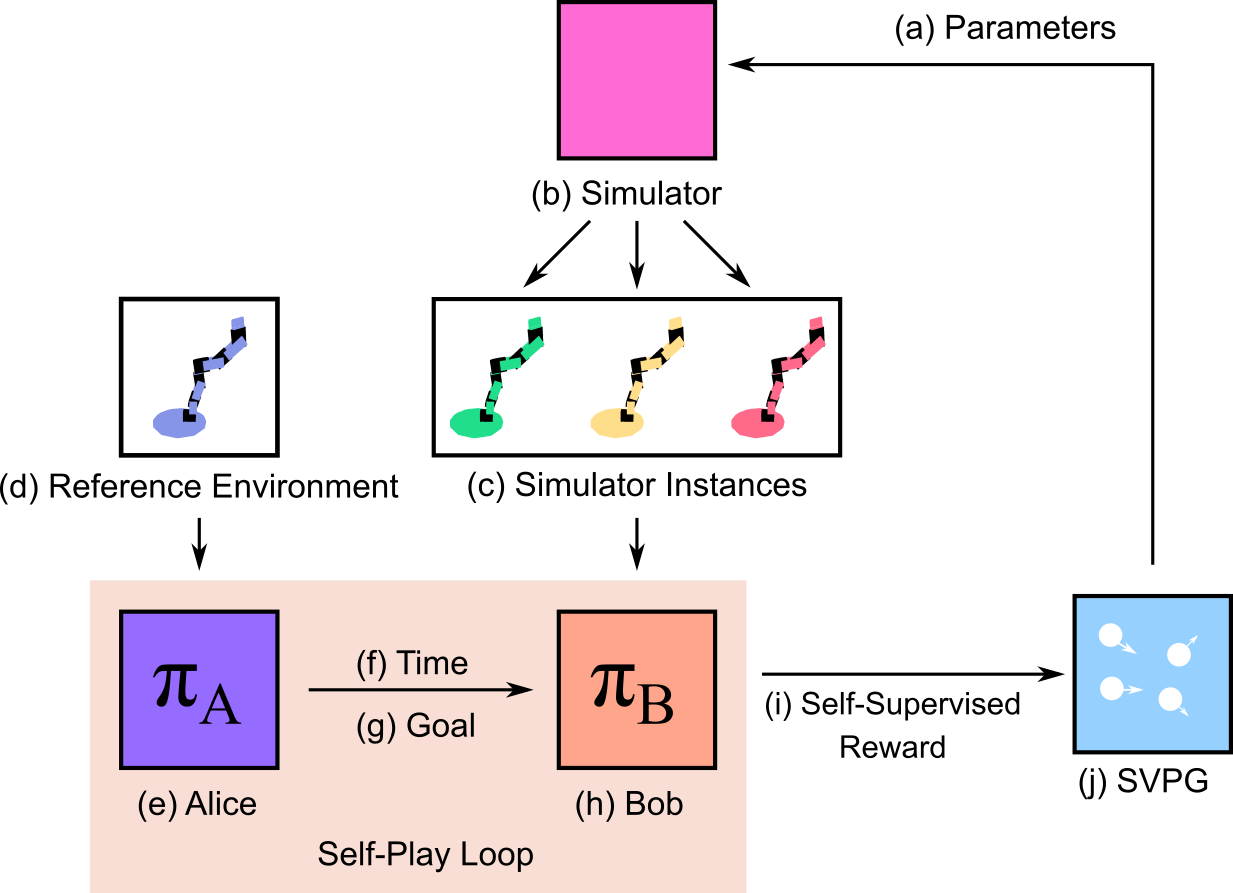}
  \end{minipage}\hfill
  \begin{minipage}[c]{0.48\textwidth}
    \captionof{figure}{Self-Supervised Active Domain Randomization (SS-ADR) learns \textbf{robust policies} (h) via self-play by co-evolving a goal curriculum, set by \textbf{Alice} (e), alongside an environment curriculum, set by the \textbf{SVPG particles} (j). The \textbf{randomized environments} (c) and \textbf{goals} (g) slowly increase in difficulty, leading to strong zero shot transfer on all environments tested.} \label{fig:ssadr}
  \end{minipage} 
\end{figure}



\section{Background}
\subsection{Reinforcement Learning}
We consider a Markov Decision Process (MDP), $\mathcal{M}$, defined by ($\mathcal{S}, \mathcal{A}, \mathcal{T}, \mathcal{R}, \gamma $), where $\mathcal{S}$ is the state space, $\mathcal{A}$ is the action space, $\mathcal{T} : \mathcal{S} \times \mathcal{A} \rightarrow \mathcal{S}$ is the transition function, $\mathcal{R}: \mathcal{S} \times \mathcal{A} \rightarrow \mathbb{R}$ is the reward function and $\gamma$ is the discount factor. Formally, the agent receives a state $s_t \in \mathcal{S}$ at the timestep $t$ and takes an action $a_t$ based on the policy $\pi_{\theta}$. The environment provides a reward of $r_t$ and the agent transitions to the next state $s_{t+1}$. The goal of \ac{RL} is to find a policy $\pi_{\theta}$ which maximizes the expected return from each state $s_t$ where the return $R_t$ is given by $R_t = \sum_{k = 0}^{\infty} \gamma^{k}r_{t+k}$. Goal-directed \ac{RL}  often appends a \textit{goal} (some $g$ in a goal space $\mathcal{G}$) to the state, and requires the goal when evaluating the reward function (i.e $\mathcal{R}: \mathcal{S} \times \mathcal{G} \times \mathcal{A} \rightarrow \mathbb{R}$).

\subsection{Self-Play}
We consider the self-play framework proposed by \citet{asymmetricselfplay}, which proposes an unsupervised way of learning to explore the environment. In this method, the agent has two brains: Alice, which sets a task, and Bob, which finishes the assigned task. This dichotomy is quantified in Equations \ref{eq:alice_reward} and \ref{eq:bob_reward}:
\begin{equation}
    r_a = \upsilon * \max(0, \ t_b - t_a)
    \label{eq:alice_reward}
\end{equation}
\vspace{-13.25pt}
\begin{equation}
   r_b = - \upsilon * t_b
    \label{eq:bob_reward}
\end{equation}
where $t_a$ is the time taken by Alice to set a task, $t_b$ is the time taken by Bob to finish the task set by Alice and $\upsilon$ is the scaling factor. This dual-reward design, $r_a$ for Alice and $r_b$ for Bob, allows self-regulating feedback between both agents, as Alice is rewarded most heavily for picking tasks that are just beyond Bob's \textit{horizon}: Alice tries to propose tasks that are easy for her, yet difficult for Bob. This evolution of tasks forces the two agents to construct a curriculum for exploration automatically. 

\subsection{Domain Randomization}
Domain randomization \citep{Sadeghi_2017, tobin2017domain} is a technique popular in robotic domain transfer, particularly in the zero-shot transfer\footnotemark\footnotetext{Zero-shot transfer does not allow an agent to take extra optimization steps in the testing distribution.} problem setup, allowing for robotic agents to train entirely in simulation. Domain randomization randomizes numerical parameters of a simulated robotic task (i.e., friction, gravity) during the training such that the agent cannot exploit the approximate dynamics within a simulator. The agent, trained on a host of randomized, simulated environments, ideally learns a policy that fits only the task - not the environment, enabling it to generalize well when it is transferred to the real robot.

Domain randomization requires the explicit definition of a set of $N_{rand}$ simulation parameters to vary, as well as their bounded domain (denoted as a \textit{randomization space } $\Xi \subset \mathbb{R} ^{N_{rand}}$). During every episode, a set of parameters $\mathbf{\xi} \in \Xi$ are sampled to generate a new MDP when passed through the simulator $S$\footnotemark\footnotetext{Domain randomization generally changes the transition function $\mathcal{T}$ by editing dynamics parameters of the simulation.}. In the standard domain randomization formulation, the parameters at each episode are sampled uniformly from the randomization space throughout training.

\subsection{Active Domain Randomization}
\ac{ADR} \citep{pmlr-v100-mehta20a} builds off the assumption that \ac{DR} leads to inefficient learning when sampling uniformly from the randomization space. \ac{ADR} poses the environment curriculum learning problem - the search over the randomization space - as a reinforcement learning problem for the most informative environment instances. In ADR, the environment sampler (the policy) is parameterized by Stein's Variational Policy Gradient (SVPG) \citep{DBLP:journals/corr/LiuRLP17}. ADR learns to control a set of particles $\{\mu_{\phi_{i}}\}_{i = 1}^{N}$ which directly correspond to which environments are shown to the agent. Each particle has its own set of parameters, $\phi_i$, which are trained with the update described in Equation \ref{eq:svpg}.

The use of SVPG allows the particles to undergo interacting updates, which includes a term which maximizes return and a term which induces diversity between particles (and correspondingly, environments shown to the robotic agent). The full update can be written as:
\begin{equation}
\begin{split}
\mu_{\phi_i} \leftarrow \mu_{\phi_i} &+ \frac{\epsilon}{N} \sum^N_{j = 1} [\nabla_{\mu_{\phi_j}}  J(\mu_{\phi_j})k(\mu_{\phi_i}, \mu_{\phi_j}) + \alpha\nabla_{\mu_{\phi_j}}k(\mu_{\phi_i}, \mu_{\phi_j})], \\
\end{split}
\label{eq:svpg}
\medskip
\end{equation}
where $J(\mu_{\phi_i}) $ denotes the sampled return from particle $i$, $k(\cdot, \cdot)$ is a kernel that calculates similarity between two particles and forces similar points in the parameter space away from each other \citep{10.5555/3157096.3157362}, and the learning rate $\epsilon$ and temperature $\alpha$ are hyperparameters. 

The particles navigate the randomization space, looking for the simulation parameters that may generate the most utility. The particles are trained by using learned discriminator-based rewards $r_D$, similar to the formulation in \citep{DBLP:journals/corr/abs-1802-06070}. The discriminator $D$ (with trainable parameters $\psi$) attempts to measure the discrepancies between the trajectories from the reference $E_{ref}$ and randomized environment instances $E_i$ (corresponding to the $i^{\text{th}}$ particle's parameter proposal $\xi_i$).

The reward coming from the discrimintator for the SVPG is defined as:

\begin{equation}
r_D = log \ D_{\psi}(y | \tau_{i} \sim \pi(\cdot;E_i))
\label{eq:discriminator-rew}
\end{equation}

where $y$ is a boolean variable denoting the source (randomized or reference) and $\tau_i$ is the randomized trajectory generated inside of the randomized instance $E_i$.

This reward formulation drives ADR to find environments which are difficult for the current agent policy to solve, as measured via learnable discrepancies between trajectories generated by the agent policy in a reference (and generally easier) environment, and a proposed randomized instance. 

\begin{figure}[tb]
    \centering
    \subfigure[]{\includegraphics[width=0.24\textwidth]{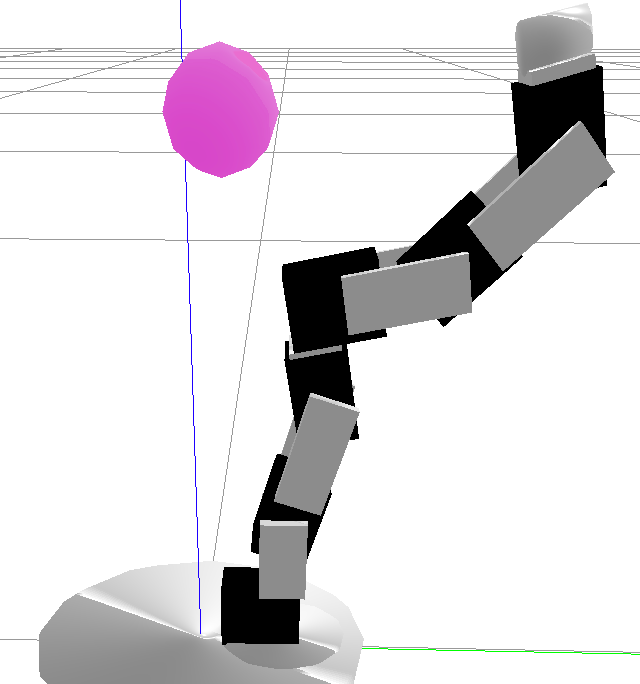}} 
    \subfigure[]{\includegraphics[width=0.24\textwidth]{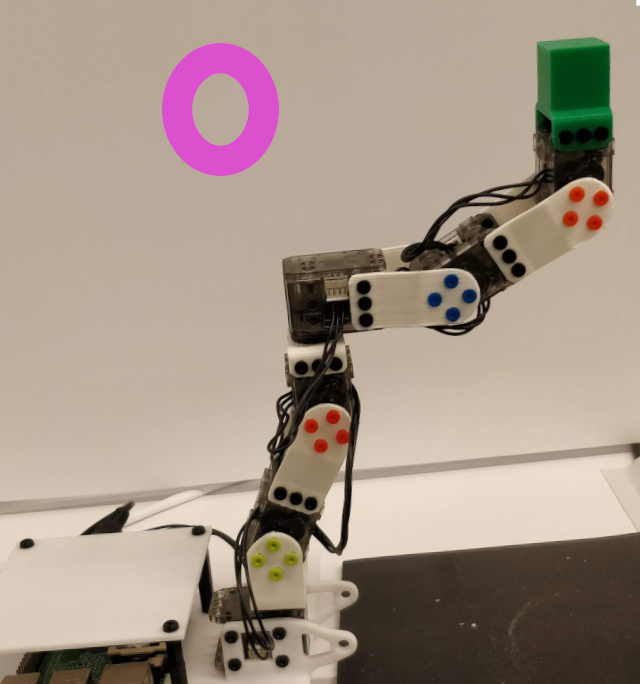}} 
    \subfigure[]{\includegraphics[width=0.24\textwidth, height=0.255\textwidth]{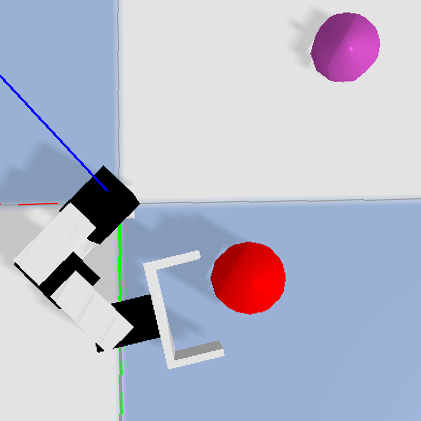}}
    \subfigure[]{\includegraphics[width=0.24\textwidth, height=0.255\textwidth]{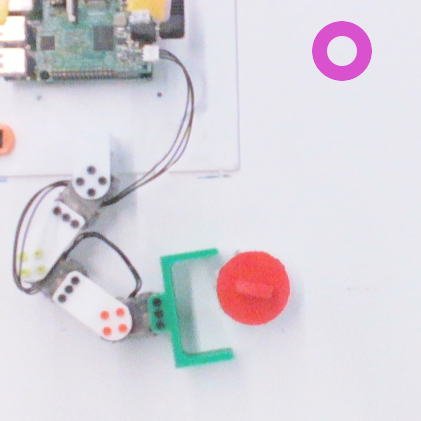}}
    \caption{(a, b) ErgoReacher is a 4 \ac{DoF} robotic arm, with both simulation and real world environments. The goal is to move the end effector to several imaginary goals (pink dot) as fast as possible, actuated with the four motors. (c, d) ErgoPusher is a 3\ac{DoF} robotic arm, with the goal of bringing a secondary object to an imaginary goal (pink dot).}
    \label{fig:robot-pics}
\end{figure}


\section{Method}


\ac{ADR} allows for curriculum learning in an environment space: given some black box agent, trajectories are used to differentiate between the difficulty of environments, regardless of the goal set in the particular environment instance. In goal-directed \ac{RL}, the goal itself may be the difference between a useful episode and a useless one. In particular, certain goals within the same environment instance may vary in difficulty; on the other hand, the same goal may vary in terms of reachability in different environments. ADR provides a curriculum in environment space, but with goal-directed environments, we have a new potential curriculum to consider: the goal curriculum.

In order to build proficient, generalizable agents, we need to evolve a curriculum in goal space \textit{alongside} a curriculum in environment space; evolving them independently may lead to degenerate solutions: the algorithm may learn to propose impossible goals with \textit{any} environment, or impossible environments (i.e., with unrealistic physical parameters) with \textit{any} goal. As shown in \citep{asymmetricselfplay}, self-play provides a way for policies to learn without environment interaction, but when used only for goal curricula, requires interleaving of self-play trajectories alongside reward-evaluated rollouts for best performance. Our work investigates the following question:

{\centering\textit{Can we co-evolve a goal and environment curriculum, both with the same self-supervised learning signal?}}

To this end, we propose \textbf{Self-Supervised Active Domain Randomization} (SS-ADR), summarized in Algorithm \ref{alg:ssadr1}. SS-ADR learns a curriculum in the joint goal-environment space using only a single self-supervised reward signal, producing strong and generalizable policies. SS-ADR can be seen in its entirety in Algorithm \ref{alg:ssadr1} and \ref{alg:ssadr2} and Figure \ref{fig:ssadr}, and is described qualitatively below.

SS-ADR learns two additional policies compared to the standard ADR formulation: \textit{Alice} and \textit{Bob}. Alice - the goal setter - operates in the environment, and eventually signals a \texttt{STOP} action. The environment is then reset to the starting state, and now Bob uses his policy to attempt to achieve the goal Alice has set. Bob sees Alice's goal state appended to the current state, while Alice sees the current state appended to it's initial state. Alice and Bob's normal policies are trained via DDPG \citep{ddpg} and environment rewards, while Alice's \texttt{STOP}-signalling policy is trained with Vanilla policy gradients using Equation \ref{eq:alice_reward}. 

To co-evolve the environment curriculum alongside the goal curriculum, we introduce the randomization aspects of our approach. Before Bob operates in the environment, the environment is \textit{randomized} (e.g. object frictions are perturbed or robot torques are changed), according to the values set from the ADR environment-sampler. Alice, who operates in the \textit{reference} environment, $E_{ref}$ (an environment given as the ``default"), tries to find goals that are easy in the reference environment ($E_{ref}$), but difficult in the randomized ones ($E_{rand}$). 

To enforce the co-evolution, we train the ADR particles with Alice's reward (i.e., Equation \ref{eq:alice_reward} is evaluated separately for each randomization prescribed by the individual ADR particles). As this reward depends on time spent by both Alice and Bob (where time for completion is denoted by $t_a$ and $t_b$ in Equation \ref{eq:alice_reward}), the curriculum in both goal and environment space evolve in difficulty simultaneously.

The reward structure forces Alice to focus on horizons: her reward is maximized when she can do something quickly that Bob cannot do at all. Considering the synchrony of policy updates for each agent, we presume that the goal set by Alice is not far out of Bob's current reach. In addition, as Bob's policy is trained with the environment reward but operates in a randomized environment, Alice's \textit{acting} policy is time-delayed copy of Bob's policy. This, \textbf{along with} the co-evolution of curriculum, greatly improves robotic transfer.
\begin{figure}[tb]
    \begin{minipage}[tb]{.49\textwidth}
      \begin{algorithm}[H]
        \captionof{algorithm}{Self Supervised ADR}
        \label{alg:ssadr1}
        \begin{algorithmic}
          \State {\textbf{Input}: $\Xi$: Randomization space, $S$: Simulator ($S: \Xi \rightarrow E$), $\xi_{ref}$: reference parameters}
          \State \textbf{Initialize} $\pi_a$: Alice's acting policy, $\pi_{a}^{s}$: Alice's stopping policy, $\pi_b$: Bob's acting policy, $\mu_{\phi}$: SVPG particles
          \For{ $T_{max}$ timesteps}
          \State $\pi_{a} \leftarrow$  Old bob's policy $\pi_b$
          \State  $E_{ref} \leftarrow S(\xi_{ref})$
          \State Observe the initial state $s_o$
          \State $t_a, t_b$ = {\fontfamily{qcr}\selectfont Self-Play}($\pi_a, \pi_{a}^{s}, \pi_b, \mu_{\phi}$)
          \State Compute Alice's reward $r_a$ using Eq (\ref{eq:alice_reward})
          \State Update $\pi_{a}^{s}$ with $r_a$ (goal curriculum) 
          \State Update the SVPG particles with $r_a$ using \\
          (\ref{eq:svpg})(environment curriculum)
          \State Update $\pi_a$ and $\pi_b$ using environment rewards
        
          \EndFor
        \end{algorithmic}
      \end{algorithm}%
    \end{minipage}%
    \hfill
    \begin{minipage}[tb]{.49\textwidth}
      \begin{algorithm}[H]
        \captionof{algorithm}{Self-Play}
        \label{alg:ssadr2}
        \begin{algorithmic}
          \State {\textbf{Input}} :  $\pi_a$, $\pi_{a}^{s}$, $\pi_b$, $\mu_{\phi}$
          \State $t_a, t_b \leftarrow 0$
          \While{$a_{t_{s}}$ is \textbf{not} \texttt{STOP}}
                \State $t_a \leftarrow t_a + 1$
                \State Observe the current state $s_{t_{a}}$
                \State $a_{t_s} \leftarrow \pi_{a}^{s}(s_o, s_{t_{a}}; E_{ref})$
                \State  $a_{t_a} \sim \ \pi_a(s_o, s_{t_{a}}; E_{ref})$
          \EndWhile
          \State Bob's target state: $s^* \leftarrow s_{t_{a}}$
          \State Sample environment $\xi_{rand} \sim \mu_{\phi}(\cdot)$ 
          \State  $E_{rand} \leftarrow S(\xi_{rand})$
          \While{Bob \textbf{not} done}
                \State $t_b \leftarrow t_b + 1$
                \State Observe the current state $s_{t_{b}}$
                \State  $a_{t_{b}} \sim \ \pi_b(s_{t_b}, s^*; E_{rand})$
                
          \EndWhile
          \State \textbf{Return: } $t_a$, $t_b$
        \end{algorithmic}
      \end{algorithm}
    \end{minipage}\par\vspace{-1\baselineskip}
\end{figure}
\section{Results}
In order to evaluate our method, we perform various experiments on continuous control robotic tasks both in simulation and real world. We used the following environments from \citep{pmlr-v87-golemo18a} and \citep{pmlr-v100-mehta20a}:

\begin{itemize}
    \item \textbf{ErgoReacher:} A 4\ac{DoF} robot where the end-effector has to reach the goal (Figure \ref{fig:robot-pics}(a, b))
    \item \textbf{ErgoPusher:} A similar 3\ac{DoF} robot that has to push the puck to the goal (Figure \ref{fig:robot-pics}(c, d))
\end{itemize}

For the \textit{sim-to-real} experiments, we recreate the simulation environment on the real Poppy Ergo Jr. robots \citep{article} shown in Figure \ref{fig:robot-pics}.  All simulation experiments are run across 4 random seeds. We evaluate the policy on \textbf{(a)} the default environment and \textbf{(b)} an intuitively hard environment which lies outside the training domain, for every 5000 timesteps, resulting in 200 evaluations in total over 1 million timesteps. 

We compare our method against two different baselines:
\begin{itemize}
    \item\textbf{Uniform Domain Randomization (UDR)}: We use UDR, which generates a multitude of tasks by uniformly sampling parameters from a given range as our first baseline. The environment space generated by UDR is unstructured and there is no intuitive curriculum. The goal stays constant throughout episodes.
    \item \textbf{Unsupervised Default}: We use the self-play framework to generate a naturally growing curriculum of goals as our second baseline. Here, only the goal curriculum (and not the coupled environment-goal curriculum) is considered. 
\end{itemize}
\subsection{Simulation Experiments}

We evaluate SS-ADR's performance on the \textit{ErgoPusher} and \textit{ErgoReacher} tasks. In the  \textit{ErgoPusher} task, we vary the puck friction ($N_{rand} = 1$). In order to create an intuitively hard environment, we lower the value of this parameter, which creates an ``icy'' surface, ensuring that the puck needs to be hit carefully to complete the difficult task.

\begin{figure}[tb]
    \centering
    \subfigure[]{\includegraphics[width=0.45\columnwidth]{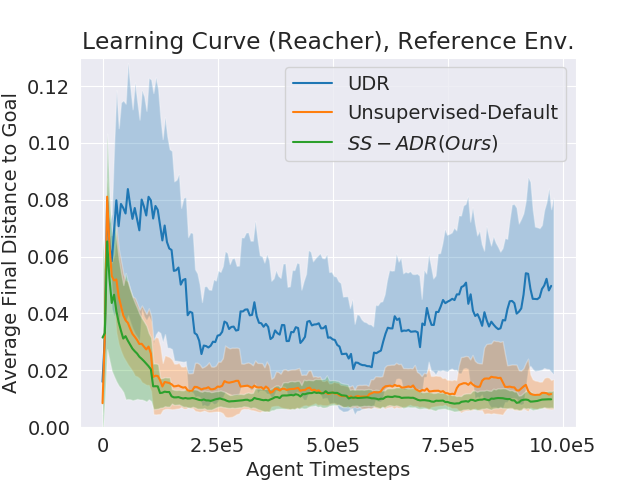}} 
    \subfigure[]{\includegraphics[width=0.45\columnwidth]{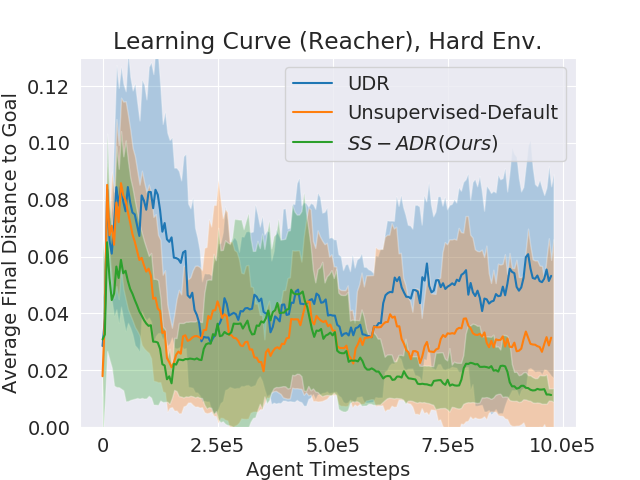}} 
    \caption{(a) On the default (in-distribution) environment, both the self-play method, shown as \textit{Unsupervised-Default}, and SS-ADR show strong performance. Even on an easier task, we see issues with UDR, which is unstable in both performance and convergence throughout training. (b) On the intuitively hard environment, we see that only SS-ADR converges with low variance and strong performance while the other baselines struggle both in terms of variance and performance. Shown is final distance to goal, lower is better.}
    \label{fig:reacher-sim}
\end{figure}

For the \textit{ErgoReacher} task, we increase the randomization dimensions ($N_{rand} = 8$) making it hard to intuitively infer the environment complexity. However, for the demonstration purposes, we create an intuitively hard environment by assigning extremely low torques and gains for each joint.  We adapt the parameter ranges from  \citeauthor{pmlr-v100-mehta20a} (\citeyear{pmlr-v100-mehta20a}).

\begin{figure}[b]
    \centering
    \subfigure[]{\includegraphics[width=0.45\columnwidth]{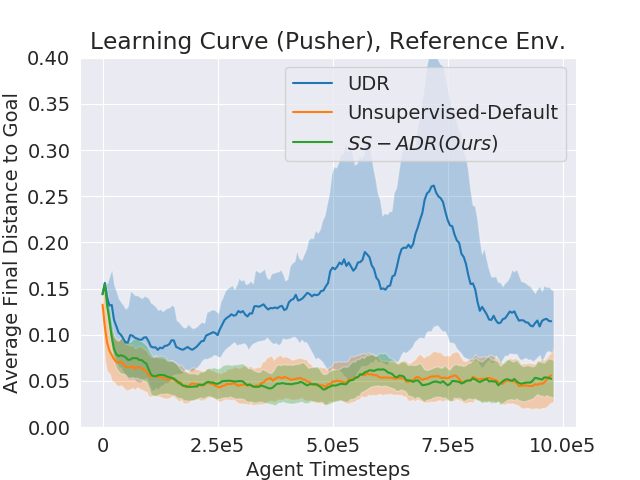}} 
    \subfigure[]{\includegraphics[width=0.45\columnwidth]{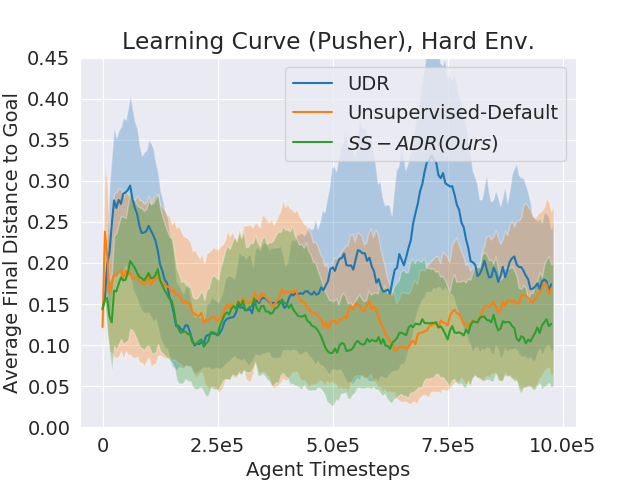}} 
    \caption{(a) In the ErgoPusher environment, we see the same narrative as in Figure \ref{fig:reacher-sim}; UDR struggles even in the easy, in-distribution environment, while both self-play methods converge quickly with low variance. (b) Both self-play methods show higher variance in an intuitively hard environment, despite the fact that SS-ADR has better overall performance. UDR, as expected, still struggles on the held out environment. Shown is final distance to goal, lower is better.}
    \label{fig:pusher-sim}
\end{figure}

From Figure \ref{fig:reacher-sim}(a) and  \ref{fig:pusher-sim}(a) we can see that both Unsupervised-Default and SS-ADR significantly outperform UDR both in terms of variance and average final distance. This highlights that the uniform sampling in UDR can lead to unpredictable and inconsistent behaviour. To actually see the benefits of the coupled environment-goal curriculum over solely goal curriculum, we evaluate on the intuitively-hard environments (outside of the training parameter distribution, as described above). From Figure \ref{fig:reacher-sim}(b) and \ref{fig:pusher-sim}(b), we can see that our method, SS-ADR, which co-evolves environment and goal curriculum, outperforms \textit{Unsupervised-Default}. This shows that the coupled curriculum enables strong generalization performance over the standard self-play formulation.

\subsection{Sim-to-Real Transfer Experiments}
In this section, we explore the zero-shot transfer performance of the trained policies in the simulator. To test our policies on real robots, we take the four independently trained policies of both \textit{ErgoReacher} and \textit{ErgoPusher} and deploy them on the real robots without any fine-tuning. We roll out each policy per seed for 25 independent trials and calculate the average final distance across these 25 trials. To evaluate the generalization, we change the task definitions (and therefore the MDPs) of the puck friction (across low, high, and standard frictions in a box pushing environment) in case of \textit{ErgoPusher} and joint torques (across a wide spectrum of choices) on \textit{ErgoReacher}. In general, lower values in both settings correspond to harder tasks, due to construction of the robot and the intrinsic difficulty of the task itself.

\begin{figure}[tb]
    \centering
    \subfigure[]{\includegraphics[width=0.45\columnwidth]{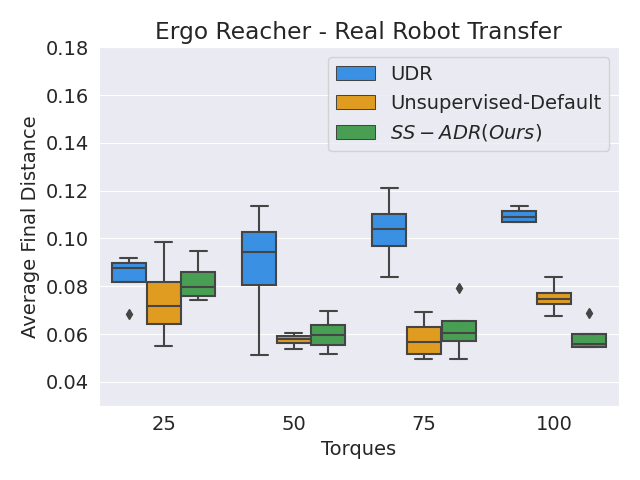}} 
    \subfigure[]{\includegraphics[width=0.45\columnwidth]{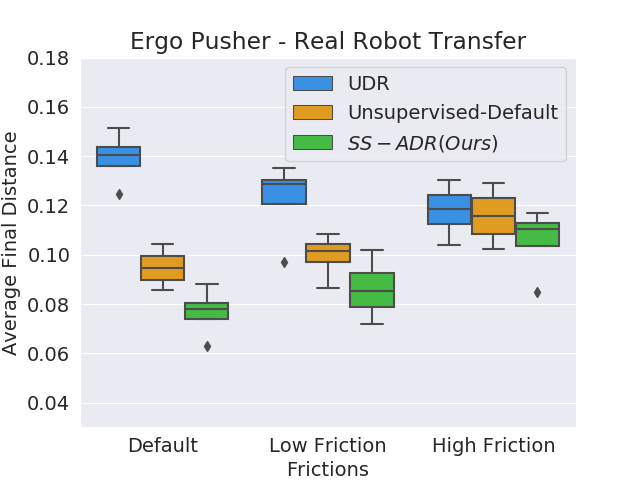}} 
    \caption{(a) On various instantiations of the real robot (parameterized by motor torques), SS-ADR outperforms UDR in terms of performance (lower is better) and spread. While SS-ADR's performance is almost consistent with or better than that of the Unsupervised-Default. (b) We see the difference between the various methods clearly in the Pusher environment, where SS-ADR outperforms all other baselines. Lower is better. }
    \label{fig:pusher-reacher-real}
\end{figure}

From the Figures \ref{fig:pusher-reacher-real}(a) and \ref{fig:pusher-reacher-real}(b), we see that SS-ADR outperforms both baselines in terms of accuracy and consistency, leading to robust performance across all environment variants tested. Zero-shot policy transfer is a difficult and dangerous task, meaning that low spread (i.e consistent performance) is required for deployed robotic \ac{RL}  agents. As we can see in the plots, simulation alone is not the answer (leading to poor performance of UDR), while self-play also fails sometimes to generate curricula  that allow for strong, generalizable policies. However, by utilizing both methods together, and \textit{co-evolving} the two curriculum spaces, we see multiplicative benefits of using curriculum learning in each separately.

\subsection{Self-calibration of SS-ADR}

In Domain Randomization, picking the ranges within which to randomize the parameters is often a trial-and-error process. These ranges often play an important role in the policy optimization and if not chosen properly, it might lead to optimization difficulties. In this section, we discuss benefits of SS-ADR as they relate to self-calibration.

\begin{figure}[tb]
    \centering
    \subfigure[]{\includegraphics[width=0.45\columnwidth]{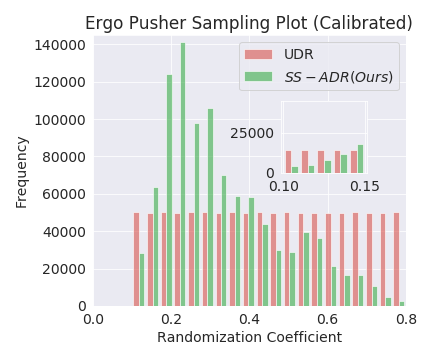}} 
    \subfigure[]{\includegraphics[width=0.45\columnwidth]{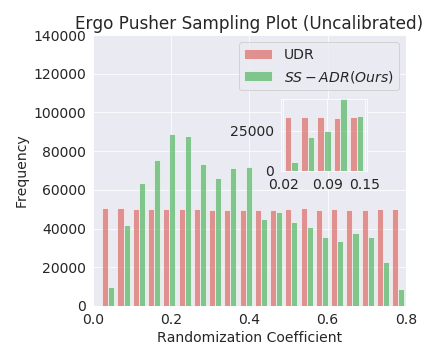}} 
    \caption{While UDR and SS-ADR both do well with calibrated ranges (a), with uncalibrated ranges, SS-ADR is the only algorithm that stays away from the physically unstable environments on the lower end of the randomization coefficient. UDR, as a static sampling algorithm, mixes samples from these environments with the rest of the range, potentially hindering training.}
    \label{fig:sampling}
\end{figure}

We train \textit{ErgoPusher} on calibrated and uncalibrated parameter ranges(which includes the impossible to solve MDPs) and obtain the environment sampling plot.  In Figure \ref{fig:sampling}(a), we see that with a calibrated range, where the ranges are carefully chosen by iteratively adjusting the bounds of the randomization space, both algorithms sample approximately equally in the ``harder" task ranges (as seen in the inset plot) although we can see SS-ADR learning the curriculum unlike UDR. In Figure \ref{fig:sampling}(b), we see the benefits of SS-ADR over UDR when using uncalibrated ranges. Here we can see that UDR is sampling certain values which generate physically unstable environments (in this experiment, approximately any environment with a randomization coefficient less than $0.05$) while SS-ADR is able to adapt and stay away from the unsolvable tasks at the extremely low end of the randomization spectrum.

\section{Related Work}
\textbf{Curriculum Learning:} The idea of curriculum leaning was first proposed by \citet{elman1993learning}, who showed that the curriculum of tasks is beneficial in language processing. Later, \citet{Bengio:2009:CL:1553374.1553380} extended this idea to various vision and language tasks which showed faster learning and better convergence. While many of these require some human specifications, recently, automatic task generation has gained interest in the \ac{RL}  community. This body of work includes automatic curriculum produced by adversarial training   \citep{DBLP:journals/corr/HeldGFA17}, reverse curriculum \citep{DBLP:journals/corr/FlorensaHWA17, DBLP:journals/corr/abs-1708-02190}, and teacher-student curriculum learning \citep{DBLP:journals/corr/MatiisenOCS17, DBLP:journals/corr/GravesBMMK17}.  However, many papers focus on distinct tasks rather than continuous task spaces and use state or reward-based ``progress" heuristics. In this work, we focus on learning the curriculum in a continuous joint task space (environments and goals) simultaneously.
 
 \textbf{Self Play: } Curriculum learning has also been studied through the lens of self-play. Self-play has been successfully applied to many games such as checkers \citep{Samuel1959SomeSI} and Go \citep{Silver_2016}.  Recently an interesting asymmetric self-play strategy has been proposed \citep{asymmetricselfplay}, which models a game between two variants of the same agent, Alice and Bob, enabling exploration of the environment without requiring any extrinsic reward.  However, in this work,  we use the self-play framework for learning a curriculum of \textit{goals}, rather than for its traditional exploration-driven use case. 
 
 
\textbf{Sim2Real Transfer:} Despite the success in deep RL, training \ac{RL}  algorithms on physical robots remains a difficult problem and is often impractical due to safety concerns. Simulators played a huge role in transferring policies to the real robot safely, and many different methods have been proposed for the same \citep{pmlr-v87-golemo18a,  DBLP:journals/corr/abs-1810-10093,  DBLP:journals/corr/abs-1810-05687}. \ac{DR} \citep{tobin2017domain} is one of the popular methods which generates a multitude of environment instances by uniformly sampling the environment parameters from a fixed range. However, \citep{pmlr-v100-mehta20a} showed that DR suffers from high variance due to an unstructured task space and instead proposed a novel algorithm that learns to sample the most informative environment instances. 
\section{Conclusion} 
In this work, we proposed Self-Supervised Active Domain Randomization (SS-ADR), which co-evolves curricula in a joint goal-environment space to create strong, robust policies that can transfer zero-shot onto real world robots. Our method solely depends on a single self-supervised reward signal through self-play to learn this joint curriculum. SS-ADR is a feasible approach to train new policies in goal-directed \ac{RL}  settings, and outperforms all baselines with low variance in both  simulated and real variants tested.
\label{sec:conclusion}

\section{Acknowledgements}
The authors gratefully acknowledge the Natural Sciences and Engineering Research Council of Canada (NSERC), the Fonds de Recherche Nature et Technologies Quebec (FQRNT), Calcul Quebec, Compute Canada,
the Canada Research Chairs, Canadian Institute for Advanced Research (CIFAR) and Nvidia for donating a DGX-1 for computation. BM would like to thank IVADO for financial support. FG would like to thank MITACS for their funding and support.


\bibliography{references}

\appendix

\section{Variance Reduction with Self-Play - ADR vs SS-ADR}

\begin{figure}[b]
    \centering
    \includegraphics[width=\textwidth]{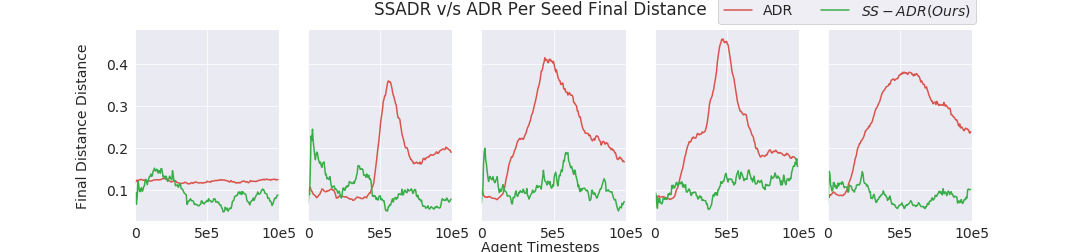}
    \caption{Differences in policy performance between ADR and SS-ADR for different fixed seeds.}
    \label{fig:per_seed_learning_curves}
\end{figure}

To investigate the stability of SS-ADR, we benchmark Active Domain Randomization in the ErgoReacher environment and plot how the policy performance (the \textit{final distance to the goal}) evolves over time. In Fig.  \ref{fig:per_seed_learning_curves}, where we can see that SS-ADR is more consistent and shows lower variance across seeds compared to ADR.

\section{Real robot comparison with ADR}

\begin{figure}[b]
    \centering
    \subfigure[]{\includegraphics[width=0.45\columnwidth]{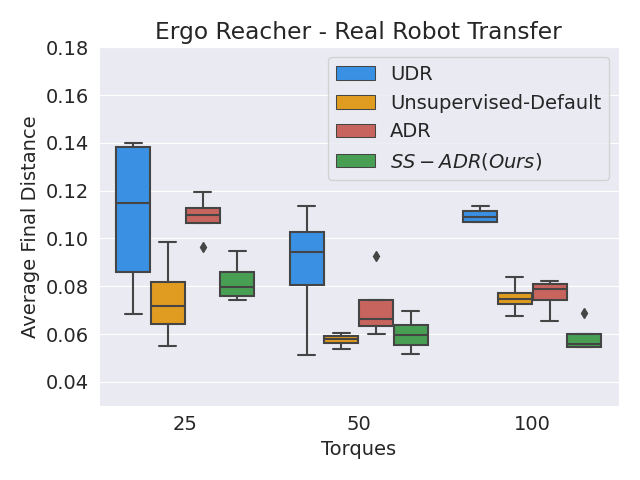}} 
    \subfigure[]{\includegraphics[width=0.45\columnwidth]{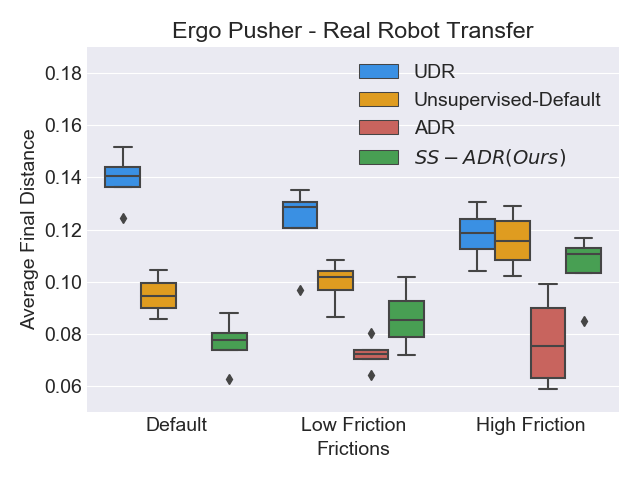}} 
    \caption{(a) On various instantiations of the real robot (parameterized by motor torques), SS-ADR outperforms UDR in terms of performance (lower is better) and spread. While SS-ADR's performance is almost consistent with or better than that of the Unsupervised-Default. (b) SS-ADR outperforms all baselines safe for ADR. }
    \label{fig:pusher-reacher-real-app}
\end{figure}

We also compared the real-robot results of SS-ADR to the results reported in \cite{pmlr-v100-mehta20a}. We could not perfectly replicate the conditions of \cite{pmlr-v100-mehta20a} as the authors used techniques similar to real robot evaluations of \cite{tobin2017domain}: \textit{low friction} consisted of manual application of lubricant to a table, and \textit{high friction} consisted wrapping an object with paper. We were unable to enter the lab to rebenchmark the two algorithms more consistently, so we report our temporary results in Fig. \ref{fig:pusher-reacher-real-app}.

\section{Implementation Details}

Across all experiments, all networks share the same network architecture and hyperparameters. For each Alice and Bob acting policy, we use Deep Deterministic Policy Gradients \citep{ddpg}, using the \texttt{OurDDPG.py} implementation from the open source repository of \citep{DBLP:journals/corr/abs-1802-09477}. Each actor and the critic have two hidden layers with 400 and 300 neurons, respectively, and use ReLU activation. For Alice's stopping policy (which signals the \texttt{STOP} action), we use a multi-layered perceptron with two hidden layers consisting of 300 neurons each. For SVPG particles we use same architecture and hyperparameters as described in \citeauthor{pmlr-v100-mehta20a} All networks use the Adam optimizer \citep{adam} with standard hyperparameters from the Pytorch implementation~\footnote{https://pytorch.org/}.The hyperparameters are summarized below:

\begin{table}[t]
  \begin{center}
    
    \label{tab:table1}
    \begin{tabular}{ c | c } 
      \textbf{Hyperparameter} & \textbf{Value}\\
      \hline
      Discount Factor $\gamma$ & 0.99 \\
      Reward scaling factor$\upsilon$ & 0.2 \\
      Actor learning rate & 0.001 \\
      Critic learning rate & 0.001 \\
      Batch size & 100 \\
      Maximum episode timesteps & 100 \\
      $N_{rand}$ for ErgoPusher & 1 \\
      $N_{rand}$ for ErgoReacher & 8 \\
    \end{tabular}
  \end{center}
\end{table}

\end{document}